\documentclass{article}
\usepackage{spconf}
\usepackage{cite}
\usepackage{amsmath,amssymb,amsfonts}
\usepackage{graphicx}
\usepackage{textcomp}
\usepackage{xcolor}
\usepackage{array}
\usepackage{float}
\usepackage[keeplastbox]{flushend}
\usepackage[inline]{enumitem}
\usepackage[utf8]{inputenc}
\usepackage[english]{babel}
\usepackage{makeidx}
\usepackage{graphicx}
\usepackage[hidelinks]{hyperref}
\usepackage{amsmath}
\usepackage[inline]{enumitem}
\usepackage{multirow}
\usepackage{subcaption}
\usepackage{caption}
\usepackage{color}
\usepackage{booktabs}
\usepackage{siunitx}
\usepackage{cite}
\usepackage{algpseudocode}
\usepackage{array}
\usepackage{float}
\usepackage{todonotes}
\usepackage[keeplastbox]{flushend}
\usepackage[linesnumbered,ruled,vlined]{algorithm2e}
\usepackage{amsfonts}
\usepackage{threeparttable}
\usepackage{ragged2e}
\usepackage[normalem]{ulem}
\DeclareUnicodeCharacter{2061}{}
\usepackage{lipsum}

\newcommand\blfootnote[1]{%
  \begingroup
  \renewcommand\thefootnote{}\footnote{#1}%
  \addtocounter{footnote}{-1}%
  \endgroup
}

\def\BibTeX{{\rm B\kern-.05em{\sc i\kern-.025em b}\kern-.08em
    T\kern-.1667em\lower.7ex\hbox{E}\kern-.125emX}}

\newcommand{\TT}[1]{{\color{black}{#1}}}

\newcommand{\RV}[1]{{\color{black}{#1}}}
\begin{document}

\title{ 
A Dilated Residual Hierarchically Fashioned Segmentation Framework for Extracting  Gleason Tissues and Grading Prostate Cancer from Whole Slide Images
}

\name{
    Taimur Hassan$^{\star}$ \qquad  
    Bilal Hassan$^{\diamond}$ \qquad
    Ayman El-Baz$^{\dagger}$  \qquad  
    \textit{Naoufel Werghi}$^{\star}$ 
}

\address{
    $^{\star}$Khalifa University, UAE,
    $^{\diamond}$ Beihang University, China,
    $^{\dagger}$University of Louisville, USA
}

\maketitle

\blfootnote{\noindent © 2021 IEEE. Personal use of this material is permitted. Permission from IEEE must be obtained for all other uses, in any current or future media, including reprinting/republishing this material for advertising or promotional purposes, creating new collective works, for resale or redistribution to servers or lists, or reuse of any copyrighted component of this work in other works.}

\begin{abstract}
\TT{Prostate cancer (PCa) is the second deadliest form of cancer in males, and it can be clinically graded by examining the structural representations of Gleason tissues. This paper proposes \RV{a new method} for segmenting the Gleason tissues \RV{(patch-wise) in order to grade PCa from the whole slide images (WSI).} Also, the proposed approach encompasses two main contributions: 1) A synergy of hybrid dilation factors and hierarchical decomposition of latent space representation for effective Gleason tissues extraction, and 2) A three-tiered loss function which can penalize different semantic segmentation models for accurately extracting the highly correlated patterns. In addition to this, the proposed framework has been extensively evaluated on a large-scale PCa dataset containing 10,516 whole slide scans (with around 71.7M patches), where it outperforms state-of-the-art schemes by 3.22\% (in terms of mean intersection-over-union) for extracting the Gleason tissues and 6.91\% (in terms of F1 score) for grading the progression of PCa.}
\end{abstract}

\section{Introduction}
\label{sec:intro}
\begin{figure}[t]
\includegraphics[width=\linewidth]{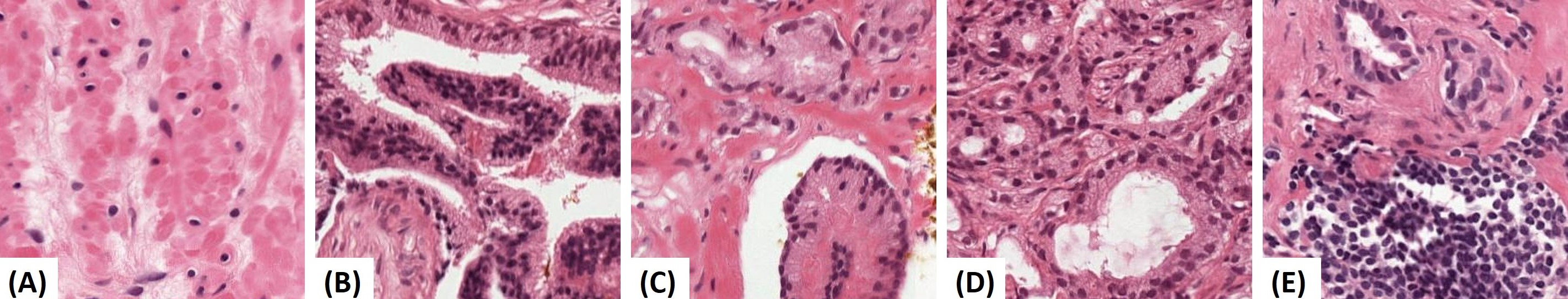}
\caption{ \small Gleason tissue patterns graded as per the ISUP grading system. (A): GrG1, (B): GrG2, (C): GrG3, (D):GrG4, and (E): GrG5.}
\centering
\label{fig:fig1}
\end{figure}

\noindent Prostate cancer (PCa) is the second most frequent form of cancer developed in men after skin cancer \cite{cancerstats, nw2}. To identify cancerous tissues, the most reliable and accurate examination is biopsy \cite{review, hassan2015Review, Smith2004JAMA}, and to grade the progression of PCa, the Gleason scores are extensively used in the clinical practice \cite{Matoso2019Histo, Wang2016EMBS}. However, in 2014, the  International Society of Urological Pathologists (ISUP) developed another simpler grading system, dubbed the Grade Groups (GrG), to monitor the PCa progression. GrG ranges from 1 to 5, where the first grade (GrG1) represents a very low risk of PCa, and GrG5 represents a severe-staged PCa. The GrG grading is performed clinically by analyzing the Gleason tissue patterns within the whole scan images (WSI), and their patches, as shown in Figure \ref{fig:fig1}. Apart from this, many researchers have diagnosed cancerous pathologies (especially PCa) from histopathology and multi-parameter magnetic resonance imagery (mp-MRI) \cite{nw1, hassan2018ICIAR2, Cao2019TMI}. The recent wave of these methods employed deep learning for segmenting the tumorous lesions \cite{Hassan2017Cancer} to grade the progression of the underlying cancer \cite{Nasim2018Gland} (specifically, the PCa \cite{Wang2018TMI, nw3, hassan2016AO}). Moreover, Gleason patterns are considered as a gold standard for identifying the severity of PCa in clinical settings \cite{Nagpal2019DM, Matoso2019Histo}. Considering this fact, Da-Silva et al. \cite{DaSilva2021JP} utilized Paige Prostate \cite{Piage}, a commercial AI-enabled cancer detection system, to classify real-world biopsy scans as normal and suspicious. Cilla et al. \cite{Cilla2021Frontiers} conducted a dosimeter study to evaluate the Pinnacle Personalized algorithm for full planning automation in prostate cancer treatments. Mun et al. \cite{Mun2021DM} developed a custom CNN model and trained it in a weakly supervised manner to screen the PCa biopsies automatically. Bulten et al. \cite{Bulten2020Lancet} also utilized deep learning to grade Gleason tissues from 5,759 locally acquired PCa biopsies. Arvaniti et al. \cite{Arvaniti2018SR} used MobileNet \cite{mobilenet} driven Class Activation Maps (CAM) for Gleason grading of the PCa tissues microarrays. In addition to this, Wang et al. \cite{Wang2017SR} conducted a study to showcase the capacity of deep learning systems for the identification of PCa (using mp-MRI) as compared to the conventional non-deep learning schemes. 

\noindent \textbf{Contributions:} \RV{Even though several frameworks have been proposed for the automatic screening of PCa based upon the Gleason scores, a robust framework, evaluated on more than 10k whole slide images (WSI), for the patch-wise extraction of Gleason tissue patterns (to grade PCa) has not been attempted yet, to the best of our knowledge.  Gleason tissues within the patched WSI are highly cluttered and correlated with each other, having similar structural and textural characteristics (see Figure \ref{fig:fig1}). The distinct characteristics within the cellular tissue structures in each patch are extremely small for the conventional segmentation models to identify them accurately. 
To address these challenges, we propose a novel encoder-decoder driven by the synergy of hybrid dilation factors \cite{hassan2019CBM, hassan2021cbm}, residual blocks \cite{hassan2019Sensors, hassan2020SoCPaR, hassan2017BOE, hassan2018Healthcom}, and hierarchical latent space decomposition \cite{hassan2020BIBE, Hassan2021JAIHC, hassan2021tim} across multiple scales to extract the diversified Gleason tissues as per the ISUP grading standards. We also propose to train this model using a multi-objective loss function to account for the class imbalance characterizing the Gleason pattern distribution. A proposed loss function also increases the capacity of conventional semantic segmentation models by many folds towards extracting the highly correlated Gleason cancerous tissues.}

\RV{\noindent The rest of the paper is organized as follows: Section \ref{sec:proposed} discusses the proposed framework, Section \ref{sec:exp} contains the experimental protocols, Section \ref{sec:results} presents the detailed evaluation of the proposed framework and its comparison with the state-of-the-art works, and Section \ref{sec:conclusion} concludes the paper and sheds light on the future directions.}

\section{Proposed Approach} \label{sec:proposed}
\begin{figure*}[t]
\centering
\includegraphics[width=1\linewidth]{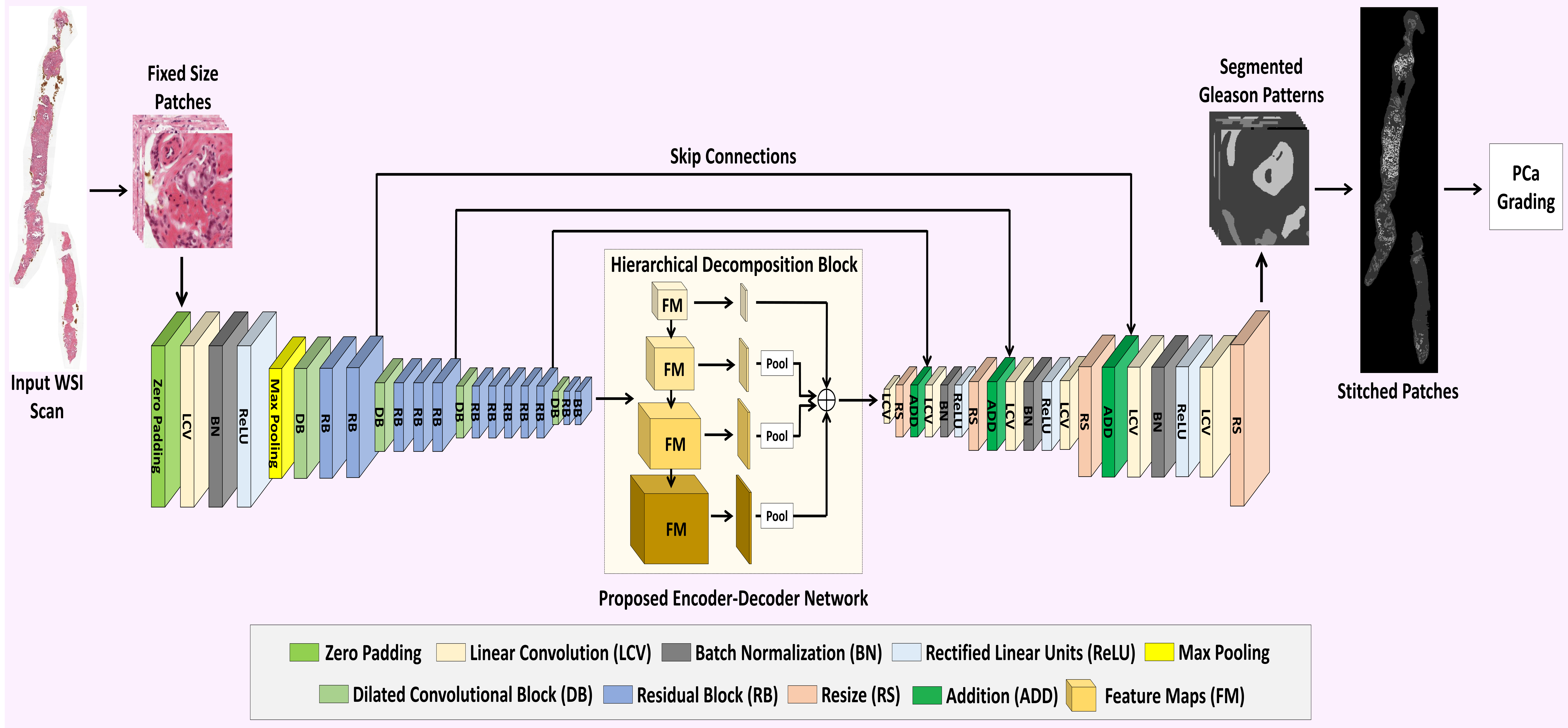}
\caption{ \small \RV{Block diagram of the proposed framework. First,  the candidate WSI is divided into fixed-size non-overlapping patches. Then, each patch is passed to the segmentation model encompassing the encoder, the hierarchical decomposition block, and the decoder. The model outcomes the different instances of the Gleason tissue patterns. These are stitched together to generate the segmented WSI representation for grading the severity of PCa.}}
\label{fig:fig2}
\end{figure*}

The block diagram of the proposed framework is shown in Figure \ref{fig:fig2}. We first divide the candidate WSI scan into a set of non-overlapping patches. These patches are then passed to a semantic segmentation model to extract the Gleason tissues as per the ISUP grading system. The extracted tissue patches are then stitched together, and the presence of the highest ISUP graded tissue is analyzed to measure the severity of PCa. Moreover,  instead of using a single loss function for training the encoder-decoder, we used a novel hybrid loss function $L_h$ that is composed of a three-tiered objective function.  A detailed description of the different framework units and the inference stage is described next.

\noindent  \textbf{The Segmentation Model:}
Cancerous tissues within the WSI patches have similar contextual and textural properties \cite{Nasim2018Gland}, making their feature representation highly correlated. Moreover, the Gleason tissue patterns, as shown in Figure \ref{fig:fig1}, exhibit a cluttered appearance at different scales. These two aspects characterizing the PCa WSI patches make the extraction and the differentiation of the Gleason tissue patterns quite challenging. To address these problematic aspects, we propose a customized semantic segmentation model. The model architecture contains an encoder, a novel hierarchical decomposition (HD) block, and the decoder. \TT{The encoder encapsulates convolutional blocks with hybrid dilation factors (DB) and the residual blocks (RB). DB performs atrous convolutions (with variable dilation factors) that increase the kernel receptive fields \cite{tbme} for accurate Gleason tissues extraction, while RB fuses the feature representation in a residual fashion to reduce computations while preserving the contextual information of the WSI patches.} The boosted latent space representation is then passed to the decoder to reconstruct the Gleason tissues, where the finer tissue details are generated from the addition driven encoder skip-connections. In addition to this, we introduce a novel HD block within the proposed framework for decomposing the extracted feature representations from the encoder block across various scales.
In so doing, we ensure better modeling and exploitation of the distinct characteristics of each Gleason tissue pattern.

\noindent \textbf{Static vs Variable Dilation Factors:} 
The dilation factors ($r$) increase the receptive field of the feature kernels for generating ampler feature representations. When $r=1$, the network performs a simple linear convolution. However, when $r>1$, the reception field of feature kernels are increased to capture more contextual information within the scans. However, employing dilated convolutional layers with the static dilation factor in a cascaded fashion might lead to the gridding effect \cite{ragnet, accv}. To overcome this, we use atrous convolutions with variable dilation factors in each stacked convolutional block. For each block (of size $n$), the dilation factors are generated through $round(r-\frac{n}{2}+i)$ (as proposed in \cite{tbme}), where $i$ varies from $0$ to $n-1$. The hyper-parameters $r$ and $n$ are determined empirically.

\noindent\textbf{Hybrid Loss Function}
\TT{We propose a hybrid loss function $L_h$ that enhance the capacity of semantic segmentation performance to recognize highly correlated cancerous tissue patterns. $L_h$ is a computed via linear combination of three-tiered objective function, as expressed below:}
\vspace{-.4cm}

\begin{equation}
    L_h = \frac{1}{N} \sum_{i=1}^N (\alpha_1 L_{c,i} + \alpha_2 L_{d,i} + \alpha_3 L_{ft,i})
    \label{eq:eq1}
\end{equation}

\vspace{-.2cm}

\noindent where 

\vspace{-.6cm}

\begin{equation}
    L_{c,i}  =  -\sum_{j=1}^{C} t_{i,j} \log⁡(p_{i,j})
    \label{eq:eq3}
\end{equation}

\vspace{-.3cm}
\begin{equation}
    L_{d,i}  =  1 - \frac{2 \sum_{j=1}^C t_{i,j} p_{i,j}} {\sum_{j=1}^C t_{i,j}^2 + \sum_{j=1}^C p_{i,j}^2}
    \label{eq:eq2}
\end{equation}

\vspace{-.6cm}

\begin{equation}
    L_{ft,i}  =  \left(1 - \frac{\sum_{j=1}^C t_{i,j} p_{i,j}} {\sum_{j=1}^C (t_{i,j}p_{i,j} + \beta_1 t_{i,j}^{'} p_{i,j} + \beta_2  t_{i,j}p_{i,j}^{'})}\right)^{1/\gamma}
    \label{eq:eq4}
\end{equation}

\noindent 
where $L_{c}$, $L_{d}$, and $L_{ft}$ denote, respectively, the categorical cross-entropy loss function \cite{hassan2016AO, hassan2016CMPB},  the dice loss function \cite{Milletari2016Dice}, and the focal Tversky loss function \cite{focalTversky2019ISBI} where $\gamma$ indicates the focusing parameter.  $t_{i,j}$ denotes the true labels of the $i^{th}$ example for the $j^{th}$ class; $p_{i,j}$ the predicted labels for the $i^{th}$ example belonging to the $j^{th}$ class; $t_{i,j}^{'}$: the true labels of the $i^{th}$ example for the non-$j^{th}$ class; and  $p_{i,j}^{'}$: the predicted labels for the $i^{th}$ example belonging to the non-$j^{th}$ class; $N$: the batch size; $C$: the number of  classes ; $\alpha_{1,2,3}$ and $\beta_{1,2}$ are the loss weights which are determined empirically. 

\noindent In conventional semantic segmentation, the networks are optimized  via standard $L_{c}$ or $L_{d}$ loss  functions.  $L_{c}$ has been attractive for its capacity to produce appealing gradients through simple subtraction between the predicted probability $p$ and the true labels.
It also achieves better convergence and is an excellent choice for the dataset having balanced classes and well-defined mask annotations \cite{icip, hassan2017AVRDB, hassan2016JOSAA, focalloss}. When the pixel-level regions, to be segmented, are scarce or imbalanced $L_{d}$ or $L_{ft}$ can be a better choice albeit at the expense of training instability when their denominators (see Eq. \ref{eq:eq2} and \ref{eq:eq4}) tend toward low values. However, $L_d$ can boost the network to achieve better overlapping regions with the ground truth, resulting in better performance (especially with imbalanced classes or ill-defined annotations). Moreover, $L_{ft}$ can ensure high resistance to imbalanced pixel-level classes, which further aids in producing better segmentation performance \cite{focalTversky2019ISBI}.

\noindent  Given the above, and considering the high correlation of the Gleason tissues and their structural and geometrical similarities, utilizing only the $L_{c}$ function can compromise the Gleason tissues extraction performance. Also, considering the scarcity of the distribution of Gleason tissues in the WSI patches, using $L_{d}$ and $L_{ft}$ alone can jeopardize the optimal convergence.  
Therefore, we hypothesize that a synergy of the three-loss functions through the proposed multi-objective function in Eq. \ref{eq:eq1} would achieve the optimal segmentation performance, accounting for the highly correlated and imbalanced cases.

\noindent \textbf{PCa Grading:}
The grading is performed WSI-wise, whereby a WSI scan is assigned as ISUP grade, the maximum GrG grade obtained in its corresponding patches \cite{isup}.
For example, if the scan patches contain GrG2, GrG3, and GrG4 tissues, then the stitched scan will be assigned a PCa severity score of GrG4.    

\section{Experimental Setup} \label{sec:exp}
\noindent  \textbf{The Dataset:}
The proposed framework has been thoroughly evaluated on a total of 10,516 multi-gigapixel whole slide images of digitized H\&E-stained biopsies acquired from 23 PCa positive subjects at the University of Louisville Hospital, USA. Each WSI scan was divided into the fixed patches of size $350 \times 350  \times 3$ (and there are around 71.7M patches in the complete dataset). Out of these 71.7M patches, 80\% of the scans were used for training, and the rest of 20\% scans are used for evaluation purposes. Moreover, all the 10,516 WSI scans contain detailed pixel-level and scan-level annotations for the ISUP grades, marked by expert pathologists from the University of Louisville School of Medicine, USA.

\noindent \textbf{Implementation:}
The implementation was conducted using TensorFlow (2.1.0) with Keras (2.3.0) on the Anaconda platform with Python (3.7.8). The training was conducted for 25 epochs with a batch size of 1024 on a machine with Intel(R) Core(TM) i9-10940X@3.30GHz CPU, 160 GB RAM NVIDIA Quadro RTX 6000 GPU with CUDA v11.0.221, and cuDNN v7.5. Moreover, the optimizer used for the training was ADADELTA \cite{Zeiler2012ADADELTA} with a learning rate of 1 and a decay rate of 0.95. The validation (after each epoch) was performed using 20\% of the training dataset. The source code has been publicly released at \url{https://github.com/taimurhassan/cancer}.

\noindent \textbf{Evaluation Metrics:} 
The segmentation performance was evaluated using the Intersection-over-Union (IoU) and the Dice Coefficient (DC). The PCa grading performance is measured scan-wise using the standard classification metrics such as true positive rate (TPR), positive predicted value (PPV), and the F1 scores.

\section{Results} \label{sec:results} 
We conducted a series of experiments that include: 1) an ablation analysis to assess the effect of the backbone network and the loss functions; 2) Comparison with the state-of-the-art semantic segmentation models for the Gleason's tissues extraction and the PCa grading.

\noindent \textbf{Effect of Backbone Network:} 
In this experiment, we evaluated how our model behaves with respect to different encoder backbones. For this purpose, we employed MobileNet \cite{mobilenet}, VGG-16 \cite{vgg16}, ResNet-50 \cite{resnet}, and the proposed Dilated Residual Network (DRN), and measured the performance of the proposed framework (employing these backbones) for extracting the Gleason tissue patterns, in terms of mean DC scores. The results, reported in Table \ref{tab:tab2}, reveals the DRN as the optimal encoder option.
\begin{table}[htb]
    \caption{Performance evaluation of the proposed framework with different backbone networks and loss functions in terms of mean DC scores.}
    \centering
    \begin{tabular}{ccccc}
    \toprule
        Backbone &  $L_c$ & $L_d$ & $L_{ft}$ & $L_h$  \\\hline
        MobileNet \cite{mobilenet} & 0.4918 & 0.5219 & 0.5059 & 0.5532  \\
        VGG-16 \cite{vgg16}   & 0.5282 & 0.5384 & 0.5554 & 0.5665  \\ 
        ResNet-50 \cite{resnet}  & 0.5414 & 0.5623 & 0.5691  & 0.5776 \\
        DRN (Proposed) & \textbf{0.5583} & \textbf{0.5694} & \textbf{0.5821} & \textbf{0.5908} \\
    \bottomrule
    \end{tabular}
    \label{tab:tab2}
\end{table}

\noindent \textbf{Effect of Loss Function:} 
In this ablation study, we experimented with the proposed model's behavior when trained with different loss functions. The results, depicted in Table \ref{tab:ls}, shows that the best Gleason patterns extraction performance is obtained with the $L_h$, confirming thus the suitability of the proposed loss function. 
 
   \begin{table}[htb]
    \caption{Effect of loss functions on the proposed framework (with DRN backbone) for extracting different Gleason tissues. Bold indicates the best score.}
    \centering
    \begin{tabular}{ccccc}
    \toprule
        Metric   & $L_c$ & $L_d$ & $L_{fh}$ & $L_h$   \\\hline
        Mean IoU & 0.3873 & 0.3891 & 0.4106 & \textbf{0.4192}  \\
    \bottomrule
    \end{tabular}
    \label{tab:ls}
\end{table}

   \begin{table}[htb]
    \caption{Gleason tissues extraction comparison in terms of  ($\mu$IoU). For fairness, all the models use proposed DRN as a backbone. The abbreviations are: LF: Loss Function, PF: Proposed Framework, DL: Dual Super-Resolution Learning \cite{dsrl}, PN: PSPNet \cite{pspnet}, UN: UNet \cite{unet}, and F8: FCN-8 \cite{fcn}.}
    \centering
    \begin{tabular}{cccccc}
    \toprule
        LF   & DL & PN & UN & F8 & PF   \\\hline
        
        $L_c$  & \underline{0.3593} & 0.3092 & 0.2401 & 0.3257 & \textbf{0.3873} \\
        
        $L_d$  & 0.3471 & \underline{0.3680} & 0.3362 & 0.3408 & \textbf{0.3891} \\
        
        $L_{fh}$  & \underline{0.3869} & 0.3784 & 0.3621 & 0.3503 & \textbf{0.4106} \\
        
        $L_h$ & \underline{0.4057} & 0.3924 & 0.3745 & 0.3591 & \textbf{0.4192}  \\
    \bottomrule
    \end{tabular}
    \label{tab:tab4}
\end{table}

\noindent \textbf{Comparison of Gleason Tissues Extraction:} 
In this experiment, we focused on the evaluation of the proposed framework's capacity for extracting the Gleason tissue patterns in comparison with the state-of-the-art models, such as DSRL \cite{dsrl}, PSPNet \cite{pspnet},  UNet \cite{unet}, and FCN-8 \cite{fcn}. We trained these competitive models with four loss functions experimented in the previous ablation study. We acted so for two reasons: 1) Ensuring fairness by using the same loss function adopted by these models (mostly the cross-entropy loss function $L_c$), and 2) assessing further the effect of the newly proposed loss function when employed with other standard models.  
\TT{The results are reported in Table \ref{tab:tab4}. First,  we notice that the best Gleason tissue extraction performance is obtained with $L_h$, across all the models. This evidences that $L_h$ gives the performance boost to the conventional semantic segmentation  models for extracting the highly correlated Gleason tissues. For example, we can see in Table \ref{tab:tab4} that for second-best DSRL network \cite{dsrl}, using $L_h$ gave 11.43\% performance improvement as compared to the standard $L_c$ loss function. Also, in Table \ref{tab:tab4}, the adequacy of the proposed loss function $L_h$ can be further evidenced towards addressing the imbalanced aspect of the Gleason tissues. Moreover, looking at the results obtained with $L_h$ (Table \ref{tab:tab4} last column), we can see that, although, the proposed framework achieves 3.22\% improvements over the second-best Dual Super-Resolution Learning (DSRL) \cite{dsrl} framework. But it should be noted that DSRL \cite{dsrl}  here has also been trained with the proposed $L_h$ loss function, which resulted in its performance boost. If we compare the proposed framework's performance with the $L_c$ trained DSRL \cite{dsrl} network, we can see that it is lagging from the proposed framework by 14.28\%, which is quite significant.}

\noindent \textbf{Comparison of PCa Grading:}
In this experiment, we compared the proposed framework's performance with the state-of-the-art schemes towards correctly classifying the severity of PCa in each WSI scans. 
The comparison is reported in Table \ref{tab:tab5} in terms of scan-level TPR, PPV, and F1 scores. Here, we can see that the proposed framework for each grade group leads the state-of-the-art frameworks in terms of PPV and F1 scores. Although it lags from the DSRL \cite{dsrl} by 2.33\% in terms of TPR for grading GrG4, nevertheless, it achieved 6.91\% improvements in terms of F1 score. Furthermore, we also want to point out the fact that, in this study, the grading performance is directly related to each network's capacity for correctly extracting the Gleason tissues. 
\begin{table}[htb]
    \caption{PCa grading comparison. For fairness, all models use proposed DRN network as a backbone. Bold indicates the best performance while the second-best scores are underlined. The abbreviations are: CC: Classification Category, PF: Proposed Framework, DL: Dual Super-Resolution Learning \cite{dsrl}, PN: PSPNet \cite{pspnet}, UN: UNet \cite{unet}, and F8: FCN-8 \cite{fcn}.}
    \centering
    \begin{tabular}{ccccccc}
    \toprule
        CC  & MC & PF  & DL & PN & UN & F8 \\\hline
        GrG1 & TPR & \textbf{0.560} & 0.493 & \underline{0.524} & 0.494 & 0.461\\
             & PPV & \textbf{0.346} & 0.284 & 0.274 & \underline{0.286} & 0.217\\
             & F1 & \textbf{0.428} & 0.361 & 0.360 & \underline{0.362} & 0.295\\\hline
             
        GrG2 & TPR & \textbf{0.723} & 0.630 & 0.598 & \underline{0.702} & 0.462\\
             & PPV & \textbf{0.564} & \underline{0.511} & 0.484 & \underline{0.511} & 0.401\\
             & F1 & \textbf{0.634} & 0.564 & 0.535 & \underline{0.592} & 0.429\\\hline
        
        GrG3 & TPR & \textbf{0.450} & \underline{0.406} & 0.389 & 0.292 & 0.390\\
             & PPV & \textbf{0.107} & \underline{0.084} & 0.076 & 0.064 & 0.064\\
             & F1 & \textbf{0.174} & \underline{0.140} & 0.127 & 0.105 & 0.110\\\hline
             
        GrG4 & TPR & \underline{0.752} & \textbf{0.770} & 0.706 & 0.727 & 0.678\\
             & PPV & \textbf{0.335} & \underline{0.300} & 0.273 & 0.294 & 0.232\\
             & F1 & \textbf{0.463} & \underline{0.431} & 0.394 & 0.418 & 0.346\\\hline
             
        GrG5 & TPR & \textbf{0.578} & \underline{0.544} & 0.460 & 0.422 & 0.437\\
             & PPV & \textbf{0.138} & \underline{0.113} & 0.093 & 0.093 & 0.075\\
             & F1 & \textbf{0.223} & \underline{0.188} & 0.155 & 0.153 & 0.128\\
    \bottomrule
    \end{tabular}
    \label{tab:tab5}
\end{table}

\begin{figure}[htb]
\includegraphics[width=\linewidth]{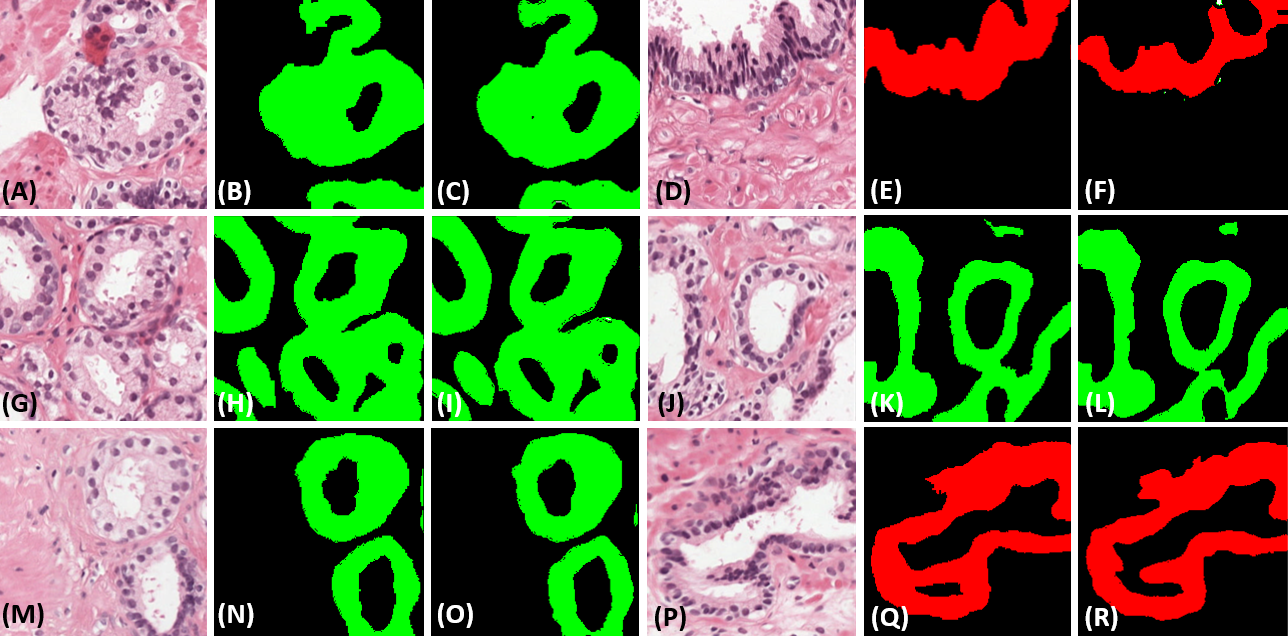}
\centerline{ \small{  (a) \hspace{0.95cm} (b) \hspace{0.92cm} (c) \hspace{0.8cm} (d) \hspace{0.98cm} (e)  \hspace{0.98cm} (f)}}
\caption{Qualitative results, (a,d) original patches, (b,e) ground truths, (c,f) the extracted tissues. Here, the red color indicates GrG2, the green color shows GrG3, and the white color shows GrG4 tissues.}
\centering
\label{fig:fig3}
\end{figure}

\noindent \textbf{Qualitative Evaluations:}
Figure \ref{fig:fig3} shows the qualitative evaluations of the proposed framework (trained with $L_h$ loss function). Here, we can see that the Gleason extraction performance is reasonable compared to the ground truth. For example, see the cases in (B)-(C), (H)-(I), and (N)-(O). Although, there are some false positives (e.g., see tiny white and green regions in F) and some false negatives (e.g., see the smaller missed region in L). But such incorrect predictions can be easily catered through morphological post-processing. \RV{More details on this are presented in the next section.}

\RV{
\noindent \textbf{Limitations:}
Since the proposed framework, like other segmentation models \cite{hassan2018JOMS, hassan2020DIR, hassan2017BioMed, hassan2017CSCI, hassan2019CCODE, hassan2018JDI, hassan2016CMPB, hassan2021tim}, is driven via pixel-wise recognition to extract similarly textured Gleason tissue patterns from the WSI patches, there is a possibility of getting false negatives (when the Gleason tissue pixels are misclassified as background pixels) and false positives (when the background pixels are misclassified as Gleason tissues). However, we also want to highlight that although the probability of having these false positives and false negatives is very less compared to other segmentation schemes (as evident from Table 3), these misclassifications can be easily cured by employing morphological schemes such as region-filling, region-opening, and blob filtering, as a post-processing step. 
}

\vspace{-0.5cm}
\section{Conclusion} \label{sec:conclusion}
This paper presents a novel encoder-decoder that leverages the hierarchical decomposition of feature representations to robustly extract Gleason tissues, which can objectively grade PCa as per the clinical standards. We have rigorously tested the proposed framework on a dataset consisting of 10,516 WSI scans. In the future, we plan to apply the proposed framework to grade other WSI-based cancerous pathologies.

\vspace{0.5cm}
\noindent \textbf{Acknowledgement}

\noindent This work is supported by a research fund from Terry Fox Foundation, Ref: I1037, and Khalifa University, Ref: CIRA-2019-047.

\small

\end{document}